# When Saliency Meets Sentiment: Understanding How Image Content Invokes Emotion and Sentiment

**Honglin Zheng**[1], **Tianlang Chen**[2], **Jiebo Luo**[3]

Department of Computer Science
University of Rochester, Rochester, NY, 14627
[1]hzheng10@u.rochester.edu, [2]t.chen@rochester.edu, [3]jluo@cs.rochester.edu

## Abstract

Sentiment analysis is crucial for extracting social signals from social media content. Due to the prevalence of images in social media, image sentiment analysis is receiving increasing attention in recent years. However, most existing systems are black-boxes that do not provide insight on how image content invokes sentiment and emotion in the viewers. Psychological studies have confirmed that salient objects in an image often invoke emotions. In this work, we investigate *more fine-grained and more comprehensive interaction between visual saliency and visual sentiment*. In particular, we partition images in several primary scene-type dimensions, including: open-closed, natural-manmade, indoor-outdoor, and face-noface. Using state of the art saliency detection algorithm and sentiment classification algorithm, we examine how the sentiment of the salient region(s) in an image relates to the overall sentiment of the image. The experiments on a representative image emotion dataset have shown interesting correlation between saliency and sentiment in different scene types and in turn shed light on the mechanism of visual sentiment evocation.

## Introduction

**Visual Sentiment Analysis**

In such an era of information explosion, it is increasingly common for people to express their emotions through posting images on social media like Twitter and Flickr. It is important for both psychologists and computer vision researchers to utilize such tremendous information to interpret the human emotion carried in the images.

In the early days, manually crafted features, e.g., pixel level features such as color, texture and composition (Machajdik and Hanbury 2010), were designed to study the emotional reaction to visual content. Similar methods were pervasive for a while until recently, when large scale image datasets such as ImageNet (Deng *et al.* 2009) and Places Dataset (Zhou *et al.* 2014) became available. The power of deep learning, especially convolutional neural networks, has been harnessed recently to discover the sentiment carried in images (You *et al*. 2015; Jindal and Singh 2015). However, most of the work are dedicated to fine-tuning pre-trained deep neural network models, such as AlexNet, VGGNet and ResNet. Most models are used as black boxes and little research has paid attention to what elements or attributes of the image are responsible for invoking emotions. In this work, we focus our effort on understanding a few dominant factors and attributes behind image sentiment evocation.

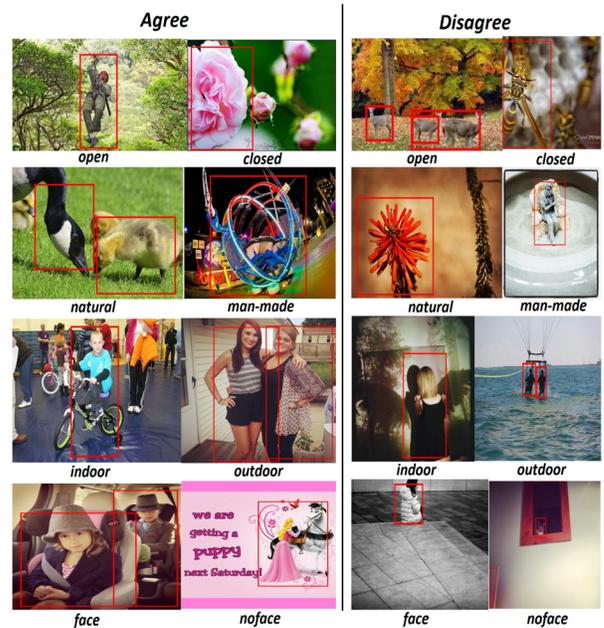

*Figure 1: Red boxes indicate detected salient objects. Left column shows example images whose salient objects share similar sentiment with the whole images. Right column row shows example images whose salient objects do not agree with the whole images in terms of sentiment. Text under each image indicates its dominant attribute over a corresponding attribute pair. Please refer to the text for detailed description of sentiment agreement and dominant attribute.*

**Saliency Detection**
In the meantime, saliency detection is also shifting from common visual feature based classifiers to deep learning based models. Previously, low-level saliency priors such as contrast prior and center prior are leveraged to approximate human observed saliency. Recently, inspired by convolutional neural network's state-of-the-art performance on various computer vision tasks such as image classification (Krizhevsky, Sutskever, and Hinton 2012) and scene recognition (Zhou *et al.* 2014), researchers start to utilize CNN to capture high-level visual concepts and produce saliency models with better detection performance (Wang *et al.* 2015; Zhang *et al.* 2015).

Saliency detection focuses on attention analysis, while visual sentiment analysis focuses on emotion analysis. Various psychologists and neuroscientists have been actively studying the underlying relationship between attention and emotion. There is an ongoing discrepancy between those who suggest that emotional perception is automatic, namely in the manner that it is independent of top-down factors such as attention (Vuilleumier *et al.* 2001), and those illustrating the dependence on attention (Grossberg, 1980; Fox *et al.* 2001). To the best of our knowledge, however, no computer vision research has been done to discover how emotion depends on attention, especially salient object(s), in an image. Even though it has been an active research topic in neuroscience, most of the neuroscientists fail to, or do not pay enough attention to the following questions:
1. Given an image, is there any salient object in it? If there is, does the salient object share similar sentiment with the entire image?
2. For those images that share similar sentiment with their salient object(s), what category do they fall into? Do they share some common attributes like man-made objects or human faces? What about those images that do not share similar sentiment with their salient objects?

Therefore, we target our work on the questions above and make several contributions:
1. We investigate fine-grained interaction between visual saliency and visual sentiment over several primary scene types, including open-closed, natural-manmade, indoor-outdoor, and face-noface.
2. We employ the state of the art saliency algorithm and visual sentiment classification model to facilitate accurate region-level computational analysis of their interactions.
3. We utilize a large public image emotion dataset to discover the relationship between saliency and sentiment over different scene types in order to understand the evocation mechanism of visual sentiment.

# Related Work

To the best of our knowledge, there is no related work on combining saliency detection and image sentiment classification to understand how image invokes human emotion. The most relevant work is image sentiment localization. Sun *et al.* (2016) proposed the notion of *AR*, Affective Region, which is a specific region in the image that contains one or more salient objects that can attract viewers' attention and carry significant emotion. In their work, an off-the-shelf object detection algorithm (Alexe, Deselaers and Ferrari 2010) is adopted to generate random proposals, and then the sentiment of both the proposals and the entire image will be evaluated to discover affective regions. However, this method utilizes explicit object recognition, which has to generate thousands of proposal windows in order to yield to a high recall rate of affective regions proposals. It also incurs a high computational overhead and may result in regions containing tiny objects that do not even catch people's attention.

In Peng *et al.* (2016)'s work, they try to predict an Emotion Stimuli Map (ESM), which describes pixels-wise contribution to emotion. They use a pre-trained and fine-tuned model to predict ESM and conclude that neither saliency nor objectness can correctly predict the image regions that evoke emotion. However, they fail to justify the choice of a dataset that consists of predominantly landscape scenes, and does not give any insight on why the saliency detection method does not work for detecting affective emotion region of such images.

Yuan *et al.* (2013) extract scene descriptor low-level features from the SUN Dataset (Xiao *et al.* 2010) and use those features to train an SVM-based classifier in order to generate 102 mid-level scene attributes, based on which they perform sentiment prediction. In Zhou *et al.* (2015)'s work, they learn deep features for images from the Places Dataset (Zhou *et al.* 2014) and are able to generate receptive field of the image for scene recognition task. The generated receptive fields reflect which part of the image the model is looking at when recognizing scenes. It is critical to understand what the scene recognition model takes into account when performing scene recognition, and similarly, it is of great significance for a sentiment analysis model to understand what scene attributes account for the prediction of image sentiment. Therefore, we propose a framework to understand for those images whose salient object(s) share the same sentiment with the entire image, what scene attributes they possess. On the other hand, for those images where salient object(s) does/do not agree with the sentiment of the entire image, what other scene attributes they possess.

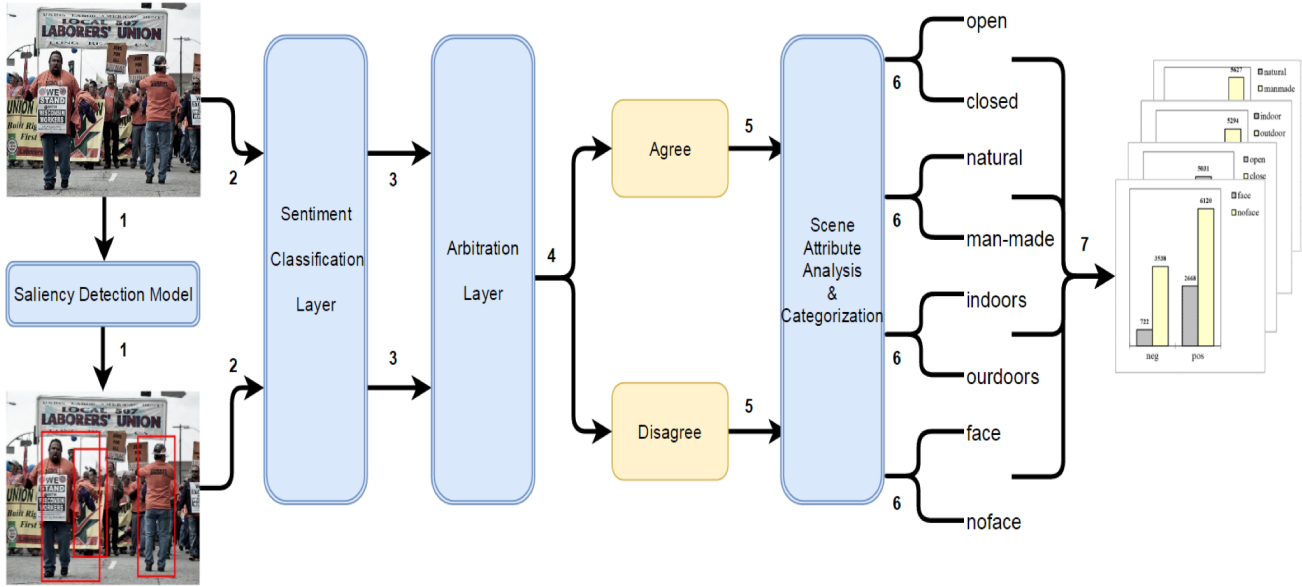

*Figure 2. Overview of our framework. Details of individual steps are described in the Methodology section.*

## Methodology

**Framework Overview**
Figure 2 illustrates our framework with annotation. Details of the steps are as follows:
*Step 1:* For a given image, we use a state-of-the-art saliency detection model (Zhang *et al.* 2016) to detect salient object(s), if any, for each image. Red boxes show the detected salient objects.
*Step 2:* We then use a state-of-the-art sentiment classification model (Campos, Jou and Giro-i-Nieto 2016) to obtain the sentiment scores for both the whole image and all of its salient objects, if any.
*Step 3 & 4:* Based on the obtained scores we are able to find out for each image, if any salient object shares the same sentiment with the entire image. Consequently, we can partition the entire dataset into two parts: those images where at least one of their salient objects shares the same sentiment with the whole image, and those images where none of their salient objects can express the same sentiment as the whole image. For simplicity, images without any salient object detected are simply excluded from further analysis. Please refer to the Visual Sentiment Analysis section for details about the definition of sentiment agreement.
*Step 5 & 6:* Within the two partitions we obtain from previous steps, we apply a state-of-the-art scene attribute detector and human face detector to further partition each part based on the detected attribute of the image.
*Step 7:* We evaluate the results of the classification mentioned above and gather any interesting findings.

**Dataset**
We use the data set released by You *et al.* (2016) for visual sentiment prediction. This dataset has 8 different categories including "*Amusement*", "*Anger*", "*Awe*", "*Contentment*", "*Disgust*", "*Excitement*", "*Fear*" and "*Sadness*". For each category we randomly sample 30% of the first 8000 images for experiment. The statistics of our current data set is shown in Table 1. Notice that this work actually does not need the exact emotion label for each image, so we can build our data set using both labeled and unlabeled images.

**Salient Object Detection**
We adopt a state-of-the-art saliency detection model (Zhang *et al.* 2016) to detect whether an image has any salient object and if there is, locate those salient objects. This model leverages the high expressiveness of the VGGNet to generate a set of scored salient object proposals for an image, based on which it produces a compact set of detected regions using a subset optimization formulation. For every image in the given dataset, the model will detect whether there is any salient object in it and save all of the detected salient objects as sub images in a corresponding folder for our subsequent scene-attribute based partitioning.

*Table 1. Statistics of the dataset.*

| # total image | # images with salient objects | # salient objects | # avg salient objects per image |
|---|---|---|---|
| 18832 | 13048 | 17920 | 1.37 |

**Visual Sentiment Analysis**
We employ Convolutional Neural Network (CNN) to obtain visual sentiment prediction. Specifically, we use the state-of-the-art Convolutional Neural Network model proposed by Campos, Jou and Giro-i-Bieto (2016). In their work, they fine-tuned a CNN model, using several performance boosting techniques to improve the performance of visual sentiment prediction. For each image, we extract the two output nodes of the last fully connected layer, and denote the sentiment probability outputs of image $i$ as $p_i = (\Pr(p)_i, \Pr(n)_i)$, which represent the probability of the emotion being negative and positive, respectively. Based on this notation we can define *SAR, Sentiment Agreement Rate*. For each image $i$,

$$SAR(i) = \min(|\Pr(p)_i - \Pr(p)_s|), \ s \in S(i)$$

where $S(i)$ is the set of all salient objects of $i$

We define that for an image, its salient object(s) *agrees* with the sentiment of the whole image if $SAR(i) \leq \theta$ where $\theta$ is the threshold. In other words, for an image, if the sentiment score of at least one of its salient object(s) is within the range of $\theta$ of the whole image's sentiment score, then this image is regarded as an image whose salient object(s) *agrees* with the sentiment of the entire image, and is classified to the partition of *agree*. Otherwise, it is an image whose salient object(s) *disagrees* with the sentiment of the entire image and is classified to the partition of *disagree*. In the experiment, we use the agreement threshold of 0.08, 0.1, 0.15, 0.18 and 0.2.

**Attribute Extraction and Categorization**
For each partition we obtain from the previous step, we fine-tune a state of the art scene recognition model, Places-CNN (Zhou *et al.* 2014), to further classify it into indoors and outdoors. This model has the same architecture as the one used in the Caffe reference network (Jia *et al.* 2014), and it is trained on the Places Database (Zhou *et al.* 2014), which is a benchmark for scene recognition. For a given image, the model will produce a 205x1 vector, and each of its dimension corresponds to the probability of one of 205 scenes. According to the given indoors/outdoors label reference, we use the labels of top 5 predicted places categories to vote whether it is indoor or outdoor. Next, we use the deep features from the last fully connected layers to detect 102 SUN scene attributes (Patterson and Hays 2012). We ranked the probability scores in a descending order. We then compare scores of *open* vs *closed*, *natural* vs *man-made*. Whichever attribute of a pair has a higher probability score than the other one is used to represent the image. We define that attribute as a *dominant attribute* in its attribute pair. For example, an image has score of 1.2 for *open* and 3.4 for *closed*, 5.6 for *natural* and 7.8 for *man-made*, then this image will be classified to have *dominant attribute* of *closed* and *manmade*. Finally, we utilize a leading cloud-based face recognition service *Face++* research tool kit (Megvii Inc.) to detect whether there is any face in a given image, based on which we can further classify the partition into images that contain face(s) and those that do not. In summary, the attribute extraction and categorization process is as follows:

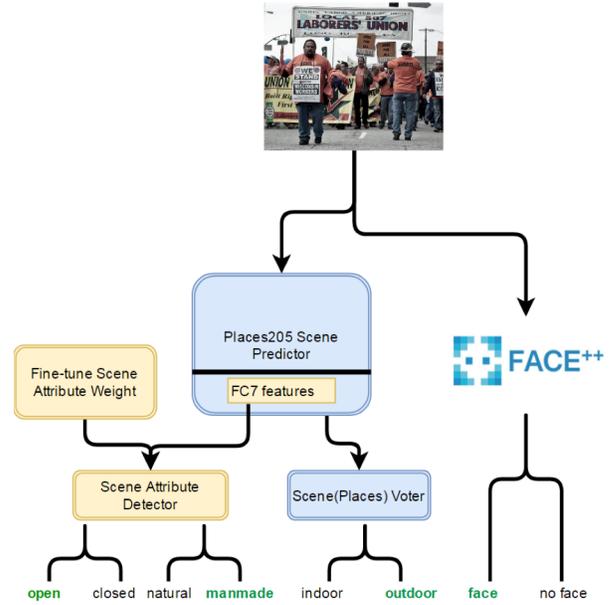

*Figure 3. Overview of attribute extraction and categorization. Bold green indicates the category the image is classified into.*

## Experiment and Evaluation

As we can see from Figure 4, the dataset is fairly well distributed among the evaluated scene attributes. With sentiment agreement threshold set to 0.15, the categorization experiment generates the results shown in Figure 5. We further conduct experiments on different sentiment agree-ment thresholds and obtain the results shown in Table 2.

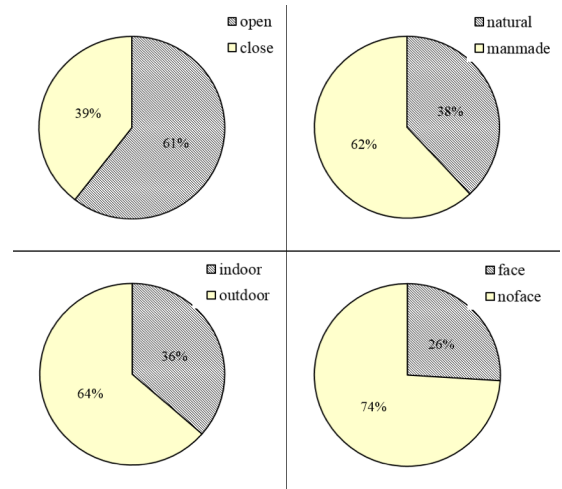

*Figure 4. Distribution of entire dataset*

To understand the table, we introduce an evaluation metric: discrimination ratio, *DR*,

$$DR(P\_A) = \frac{(\frac{\#P\_A}{\#All\_A} - \frac{\#P}{\#All})}{\frac{\#P}{\#All}}$$

where
$P(Partition) \in \{agree, disagree\}$
$A(Attribute) \in \{open, closed, natural, manmade, indoor, outdoor, face, noface\}$

#P_A: in partition P, the number of images that have dominant attribute A over its opposite. (open-closed, natural-manmade, indoor-outdoor, face-noface are opposite of each other)
#All_A: for the entire dataset, the number of images that have dominant attribute A
#P: number of images in partition P
#All: number of images in the entire dataset

If images are randomly split into two partitions of negative and positive, then the discrimination ratio for each attribute in each partition should be roughly equal to the percentage of such partition over the entire dataset. A more

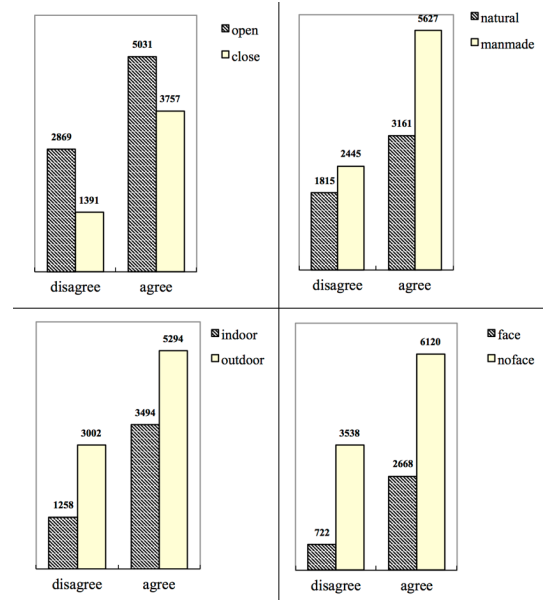

*Figure 5. Distribution of four scene attributes under the sentiment agreement threshold of 0.15.*

*Table 2. Results of scene attribute categorization with different sentiment agreement thresholds.*

| θ | disagree | open-closed | | natural-manmade | | indoor-outdoor | | face-noface | |
|---|---|---|---|---|---|---|---|---|---|
| | | dis_open / all_open | dis_closed / all_closed | dis_natural / all_natural | dis_manmade / all_manmade | dis_indoor / all_indoor | dis_outdoor / all_outdoor | dis_face / all_face | dis_noface / all_noface |
| 0.2 | 26.83 | 29.83 (+11.17) | 22.25 (-17.09) | 29.45 (+9.7) | 25.23 (-5.9) | 21.61 (-19.5) | 29.81 (+11.11) | 15.87 (-40.84) | 30.6 (+14.05) |
| 0.18 | 29.10 | 32.14 (+10.46) | 24.44 (-16.02) | 31.71 (+8.96) | 27.50 (-5.50) | 23.97 (-17.63) | 32.02 (+10.03) | 17.48 (-39.93) | 33.14 (+13.87) |
| 0.15 | 32.65 | 36.32 (+11.23) | 27.02 (-17.24) | 36.48 (+11.72) | 30.29 (-7.22) | 26.47 (-18.92) | 36.19 (+10.83) | 21.3 (-34.77) | 36.63 (+12.2) |
| 0.1 | 41.05 | 44.61 (+8.68) | 35.63 (-13.2) | 43.98 (+7.14) | 39.27 (-4.43) | 34.51 (-15.93) | 44.81 (+9.16) | 28.05 (-31.67) | 45.59 (+11.07) |
| 0.08 | 45.21 | 48.96 (+8.30) | 39.41 (-12.83) | 48.24 (+6.71) | 43.34 (-4.15) | 38.35 (-15.16) | 49.10 (+8.61) | 31.13 (-31.15) | 50.08 (+10.77) |

(a)

| θ | agree | open-closed | | natural-manmade | | indoor-outdoor | | face-noface | |
|---|---|---|---|---|---|---|---|---|---|
| | | agr_open / all_open | agr_closed / all_closed | agr_natural / all_natural | agr_manmade / all_manmade | agr_indoor / all_indoor | agr_outdoor / all_outdoor | agr_face / all_face | agr_noface / all_noface |
| 0.2 | 73.17 | 70.17 (-4.09) | 77.75 (+6.27) | 70.55 (-3.57) | 74.77 (+2.19) | 78.39 (+7.14) | 70.19 (-4.08) | 84.13 (+14.98) | 69.4 (-5.15) |
| 0.18 | 70.90 | 67.86 (-4.29) | 75.56 (+6.58) | 68.29 (-3.68) | 72.50 (+2.26) | 76.03 (+7.24) | 67.98 (+4.12) | 82.52 (+16.39) | 66.86 (-5.69) |
| 0.15 | 67.35 | 63.68 (-5.45) | 72.98 (+8.36) | 63.52 (-5.68) | 69.71 (+3.5) | 73.53 (+9.17) | 63.81 (-5.25) | 78.7 (+16.85) | 63.37 (-5.92) |
| 0.1 | 58.95 | 55.39 (-6.05) | 64.37 (+9.19) | 56.02 (-4.97) | 60.73 (+3.02) | 65.49 (+11.09) | 55.19 (-6.38) | 71.95 (+22.05) | 54.41 (-7.71) |
| 0.08 | 54.79 | 51.04 (-6.85) | 60.59 (+10.58) | 51.76 (-5.54) | 56.66 (+3.42) | 61.65 (+12.51) | 50.90 (-7.11) | 68.87 (+25.71) | 49.92 (-8.88) |

(b)

\* For both charts, all numbers except for column 1 are in percentage.
\* Number in parenthesis are *DR* rate for the correspondant attribute and partition.
\* θ: agreement threshold.
\* disagree: percentage of images whose $SAR > \theta$.
\* agree: percentage of images whose $SAR \leq \theta$.
\* dis(agr)_open(closed, natural, manmade, etc.): please refer to definition of *DR*.

positive discrimination ratio of an attribute *A* in a partition *P* indicates that an image with such an attribute is easier to be classified into *P*. In other words, the higher the discrimination ratio is, the more likely the images in partition *P* are going to have attribute *A,* and vice versa. Table 2 suggests some similar patterns for various sentiment agreement thresholds ranging from 0.08 to 0.2:

1) As the agreement threshold decreases, the *agree* partition decreases while the *disagree* partition increases, which follows the intuition that the stricter the rule is, the more difficult for a specific part of an image to represent the whole image, and it would be more like a random even split for the dataset.

2) **For images whose salient object(s) *agree* with the whole images, they tend to contain *face(s)* or outstanding *man-made* object(s), or tend to be more *closed* or more likely to be an *indoor* scene than images whose salient objects *disagree* with the whole images.**

3) From the *face-noface* column, we can see that these two attributes are influential in pulling images away from each other. With an average discrimination ratio of -35.76% in partition *disagree* and +17.96% in partition *agree*, images containing faces are highly likely to express the same sentiment as the faces express, while images without faces are more unlikely to invoke human emotion solely due to the salient objects. This observation also follows the intuition that faces tend to dominate the sentiment perception by humans. For example, we can easily tell the sentiment of the left images in Figure 1 simply by looking at the human face(s) without paying attention to other objects or elements.

We conduct another experiment to further evaluate the partitioning power of attribute combinations of *closed*, *manmade*, *outdoor* and *noface,* which have great tendency of partitioning the images to *agree*. In this experiment, we partition the *agree* and *disagree* classes according to all the possible combinations of these four pairs of attributes. For example, for each class, we calculate the total number of *closed*, *manmade*, *indoor* and *face* images, and also the total number of its entire opposite, namely *open*, *natural*, *outdoor* and *noface* images. For each combination in each partition, we will also calculate its *DR* metric.

The observations from Table 3 lead to the following interesting findings:

1) The combination with the most negative *DR* is 1, 1, 0 and 0. That is to say, **an image with an attribute combination of *open*, *natural*, *outdoor* and *noface* is more likely to be classified to the partition of *disagree*.**

2) Combinations with the face attribute are all with high *DR*, which further confirms that **images with faces will be more likely to be classified into the partition of *agree*.**

3) Surprisingly, the opposite of the most negative *DR* is not necessarily the combination with the most positive *DR*. Instead, *{closed, natural, outdoor, face}* and *{open, natural, indoor, face}* share the most positive *DR*.

*Table 3. Discrimination ratio of all pair-wise combinations of four attributes in the partition of agree.*

| open /closed | natural /manmade | indoor /outdoor | face /noface | DR |
|---|---|---|---|---|
| 0 | 0 | 0 | 0 | -3.79 |
| 1 | 0 | 0 | 0 | -10.53 |
| 0 | 1 | 0 | 0 | +1.59 |
| 0 | 0 | 1 | 0 | +3.52 |
| 0 | 0 | 0 | 1 | +17.60 |
| **1** | **1** | **0** | **0** | **-11.71** |
| 1 | 0 | 1 | 0 | +7.74 |
| 1 | 0 | 0 | 1 | +17.18 |
| 0 | 1 | 1 | 0 | +9.08 |
| **0** | **1** | **0** | **1** | **+48.48** |
| 0 | 0 | 1 | 1 | +19.12 |
| 1 | 1 | 1 | 0 | -2.72 |
| 1 | 1 | 0 | 1 | +12.58 |
| 1 | 0 | 1 | 1 | +18.23 |
| 0 | 1 | 1 | 1 | +34.98 |
| **1** | **1** | **1** | **1** | **+48.48** |

\* Threshold $\theta = 0.15$.
\* For each cell, 1 stands for the first attribute of the pair and 0 stands for the other one.
\* The most negative and positive DR rate are highlighted in bold.

## Conclusion and Future Work

In this work, using state of the art saliency detection model and visual sentiment classification model, we obtain salient object proposals and analyze their sentiment agreement with the whole images. We extract scene attributes from the images and examine the fine-grained interaction between visual saliency and visual sentiment. Our results suggest that images that contain outstanding man-made objects or human faces, or are indoors and closed, tend to express sentiment through their salient objects. On the other hand, images in which natural objects are more outstanding than man-made objects or do not contain human faces, or are outdoors and open, usually do not convey their sentiment information solely through their salient objects. We are encouraged by this study to evaluate the expressiveness of salient object(s) in terms of image sentiment. Moreover, we will conduct further experiments on the scene attribute distribution for each emotion class, e.g. *amusement* and *anger*, to study the evocation mechanism for each specific emotion. We will also study more scene attributes. It has the potential to give us more insight into what kind of attributes are responsible for invoking human emotion, providing guidance for both in-depth image sentiment analysis and psycho-visual understadning.